\documentclass[10pt,twocolumn,letterpaper]{article}
\pdfoutput=1

\usepackage{cvpr}
\usepackage{epsfig}
\usepackage{comment}
\usepackage{booktabs}
\usepackage{times,graphicx,tabulary,multirow,xspace,xcolor}
\usepackage{amsmath,amssymb,soul,setspace,pifont}
\usepackage{makecell}
\usepackage{lipsum}

\usepackage{esvect} 
\usepackage[caption=false]{subfig}

\cvprfinalcopy 

\ifcvprfinal\fi

\usepackage[pagebackref=true,breaklinks=true,letterpaper=true,colorlinks,bookmarks=false]{hyperref}

\newcommand\blfootnote[1]{%
  \begingroup
  \renewcommand\thefootnote{}\footnote{#1}%
  \addtocounter{footnote}{-1}%
  \endgroup
}

\newcommand\unhighlight[1]{\textcolor{gray}{#1}}
\newcommand{\votenet}{\textsc{VoteNet}\xspace}
\newcommand{\imvotenet}{\textsc{ImVoteNet}\xspace}
\newcommand{\ours}{\imvotenet}
\newcommand{\rgbd}{\texttt{RGB-D}\xspace}
\newcommand{\rgb}{\texttt{RGB}\xspace}

\newcommand{\cmark}{\color{gray}\ding{51}}%
\newcommand{\xmark}{\color{gray}\ding{55}}%
\newcommand{\roi}{\texttt{RoI}\xspace}
\newcolumntype{x}[1]{>{\centering\arraybackslash}p{#1pt}}
\newlength\savewidth\newcommand\shline{\noalign{\global\savewidth\arrayrulewidth
  \global\arrayrulewidth 1pt}\hline\noalign{\global\arrayrulewidth\savewidth}}
\newcommand\hshline{\noalign{\global\savewidth\arrayrulewidth
  \global\arrayrulewidth 0.6pt}\hline\noalign{\global\arrayrulewidth\savewidth}}
\newcommand{\tablestyle}[2]{\setlength{\tabcolsep}{#1}\renewcommand{\arraystretch}{#2}\centering\footnotesize}
\makeatletter\renewcommand\paragraph{\@startsection{paragraph}{4}{\z@}
  {.5em \@plus1ex \@minus.2ex}{-.5em}{\normalfont\normalsize\bfseries}}\makeatother

\def\x{\times\xspace}
\newcommand{\app}{\raise.17ex\hbox{$\scriptstyle\sim$}}

\def\cvprPaperID{5567} 

\begin{document}

\title{ImVoteNet: Boosting 3D Object Detection in Point Clouds with Image Votes}


\author{Charles R. Qi$^{*\dag}$\qquad Xinlei Chen$^{*1}$\qquad Or Litany$^{1,2}$\qquad Leonidas J. Guibas$^{1,2}$\\
$^1$Facebook AI\qquad $^2$Stanford University}


\maketitle

\begin{abstract}
3D object detection has seen quick progress thanks to advances in deep learning on point clouds. A few recent works have even shown state-of-the-art performance with just point clouds input (e.g. \votenet). However, point cloud data have inherent limitations. They are sparse, lack color information and often suffer from sensor noise. Images, on the other hand, have high resolution and rich texture. Thus they can complement the 3D geometry provided by point clouds. Yet how to effectively use image information to assist point cloud based detection is still an open question.
In this work, we build on top of \votenet and propose a 3D detection architecture called \imvotenet specialized for \rgbd scenes. \imvotenet is based on fusing 2D votes in images and 3D votes in point clouds. Compared to prior work on multi-modal detection, we explicitly extract both geometric and semantic features from the 2D images. We leverage camera parameters to lift these features to 3D. To improve the synergy of 2D-3D feature fusion, we also propose a multi-tower training scheme. We validate our model on the challenging SUN RGB-D dataset, advancing state-of-the-art results by {\bf 5.7} mAP.
We also provide rich ablation studies to analyze the contribution of each design choice. 
\end{abstract}

\section{Introduction}
\blfootnote{*: equal contributions.}
\blfootnote{\dag: work done while at Facebook.}

Recognition and localization of objects in a 3D environment is an important first step towards full scene understanding. Even such low dimensional scene representation can serve applications like autonomous navigation and augmented reality. Recently, with advances in deep networks for point cloud data, several works~\cite{voteNet,zhou2018voxelnet,shi2018pointrcnn} have shown state-of-the-art 3D detection results with point cloud as the \emph{only} input. Among them, the recently proposed \votenet~\cite{voteNet} work by Qi \etal, taking 3D geometry input only, showed remarkable improvement for indoor object recognition compared with previous works that exploit all \rgbd channels. This leads to an interesting research question: Is 3D geometry data (point clouds) sufficient for 3D detection, or is there any way \rgb images can further boost current detectors?

\begin{figure}[t!]
    \centering
    \includegraphics[width=\linewidth]{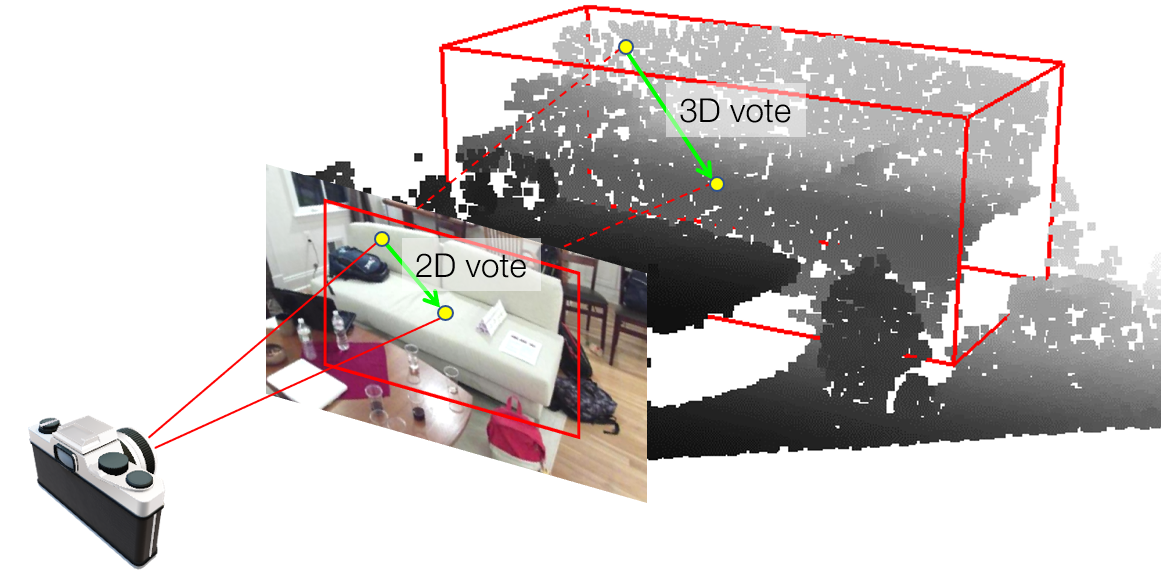}
    \caption{\textbf{Voting using both an image and a point cloud from an indoor scene.} The 2D vote reduces the search space of the 3D object center to a ray while the color texture in image provides a strong semantic prior. Motivated by the observation, our model lifts the 2D vote to 3D to boost 3D detection performance.}
    \label{fig:teaser}
\end{figure}

By examining the properties of point cloud data and \rgb image data (see for example Fig.~\ref{fig:teaser}), we believe the answer is clear:  \rgb images have value in 3D object detection. In fact, images and point clouds provide \emph{complementary} information. \rgb images have higher resolution than depth images or LiDAR point clouds and contain rich textures that are not available in the point domain. Additionally, images can cover ``blind regions'' of active depth sensors which often occur due to reflective surfaces.  On the other hand, images are limited in the 3D detection task as they lack absolute measures of object depth and scale, which are exactly what 3D point clouds can provide. These observations, strengthen our intuition that images can help point cloud-based 3D detection.

However, how to make effective use of 2D images in a 3D detection pipeline is still an open problem. A na\"ive way is to directly append raw \rgb values to the point clouds -- since the point-pixel correspondence can be established through projection. But since 3D points are much sparser, in doing so we will lose the dense patterns from the image domain.
In light of this, more advanced ways to fuse 2D and 3D data have been proposed recently. 
One line of work~\cite{qi2018frustum,xu2018pointfusion,lahoud20172d} uses mature 2D detectors to provide initial proposals in the form of frustums. This limits the 3D search space for estimating 3D bounding boxes. However, due to its \emph{cascaded} design, it does not leverage 3D point clouds in the initial detection. In particular, if an object is missed in 2D, it will be missed in 3D as well.
Another line of work~\cite{song2016deep,ku2018joint,wang2019densefusion,hou20193d} takes a more 3D-focused way to concatenate intermediate ConvNet features from 2D images to 3D voxels or points to enrich 3D features, before they are used for object proposal and box regression. The downside of such systems is that they do not use 2D images directly for localization, which can provide helpful guidance for detection objects in 3D.

In our work, we build upon the successful \votenet architecture~\cite{voteNet} and design a \emph{joint} 2D-3D voting scheme for 3D object detection named \imvotenet. It takes advantage of the more mature 2D detectors~\cite{ren2015faster} but at the same time still reserves the ability to propose objects from the full point cloud itself -- combining the best of both lines of work while avoiding the drawbacks of each.
A key motivation for our design is to leverage both geometric and semantic/texture cues in 2D images (Fig.~\ref{fig:teaser}). The geometric cues come from accurate 2D bounding boxes in images, such as the output by a 2D detector. Instead of solely relying on the 2D detection for object proposal~\cite{qi2018frustum}, we defer the proposal process to 3D. Given a 2D box, we generate 2D votes on the image space, where each vote connects from the object pixel to the 2D amodal box center. To pass the 2D votes to 3D, we \emph{lift} them by applying geometric transformations based on the camera intrinsic and pixel depth, so as to generate ``pseudo'' 3D votes. These pseudo 3D votes become extra features appended to seed points in 3D for object proposals. Besides geometric cues from the 2D votes, each pixel also passes semantic and texture cues to the 3D points, as either features extracted per-region, or ones extracted per-pixel.

After lifting and passing all the features from the images to 3D, we concatenate them with the 3D point features from a point cloud backbone network~\cite{qi2017pointnet,qi2017pointnetplusplus}. Next, following the \votenet pipeline, those points with the fused 2D and 3D features generate 3D Hough votes~\cite{hough1959machine} -- not limited by 2D boxes -- toward object centers and aggregate the votes to produce the final object detections in 3D. As the seed features have both 2D and 3D information, they are intuitively more informative for recovering heavily truncated objects or objects with few points, as well as more confident in distinguishing geometrically similar objects.

In addition, we recognize that when fusing 2D and 3D sources, one has to carefully balance the information from two modalities to avoid one being dominated by the other. To this end, we further introduce a multi-towered network structure with gradient blending~\cite{wang2019makes} to ensure our network makes the best use of both the 2D and 3D features. During testing, only the main tower that operates on the joint 2D-3D features are used, minimizing the sacrifice on efficiency. 

We evaluate \imvotenet on the challenging SUN RGB-D dataset~\cite{song2015sun}.
Our model achieves the state-of-the-art results while showing a significant improvement ({\bf +5.7} mAP) over the 3D geometry only \votenet, validating the usefulness of image votes and 2D features. We also provide extensive ablation studies to demonstrate the importance of each individual component. 
Finally, we also explore the potential of using color to compensate for \emph{sparsity} in depth points, especially for the case of lower quality depth sensors or for cases where depth is estimated from a moving monocular camera (SLAM), showing potential of our method to more broader use cases.

To summarize, the contributions of our work are:
\begin{enumerate}
    \item A geometrically principled way to fuse 2D object detection cues into a point cloud based 3D detection pipeline. 
    \item The designed deep network \imvotenet achieves state-of-the-art 3D object detection performance on SUN RGB-D.
    \item Extensive analysis and visualization to understand various design choices of the system.
\end{enumerate}

\section{Related Work}
\begin{figure*}[ht!]
    \centering
    \includegraphics[width=0.95\linewidth]{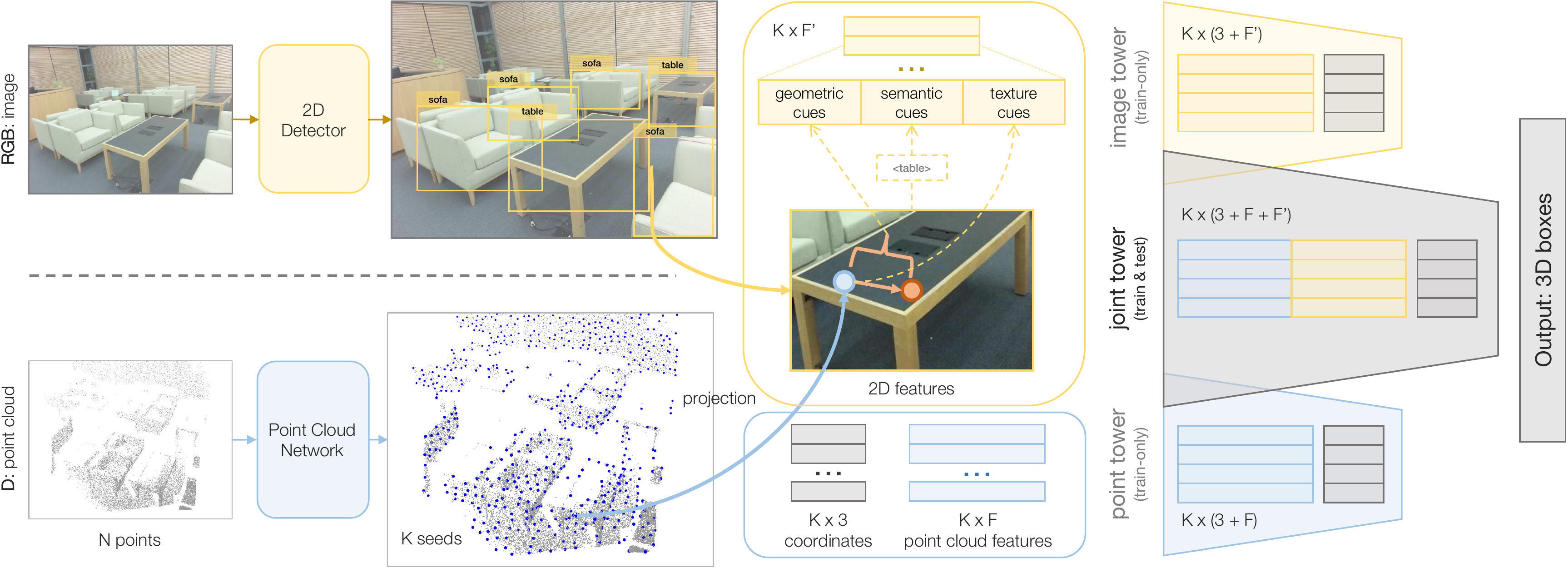}
    \vspace{2mm}
    \caption{\textbf{3D object detection pipeline for \imvotenet.}
    Given \rgbd input (with the depth image converted to a point cloud), the model initially have two separate branches: one for 2D object detection on the image and the other for point cloud feature extraction (with a PointNet++~\cite{qi2017pointnetplusplus} backbone) on the point clouds. Then we lift 2D image votes as well as semantic and texture cues to the 3D seed points (the fusion part). The seed points with concatenated image and point cloud features then generate votes towards 3D object centers and also propose 3D bounding boxes with its features (the joint tower). To push for more effective multi-modality fusion, we have two other towers that take image features only (image tower) and point cloud features only (point tower) for voting and box proposals.}
    \label{fig:pipeline}
\end{figure*}

Advances in 3D sensing devices have led to a surge of methods designed to identify and localize objects in a 3D scene.
The most relevant lines of work are detection with point clouds and detection with full \rgbd data. We also briefly discuss a few additional relevant works in the area of multi-modal data fusion.

\paragraph{3D object detection with point clouds.}
To locate objects using purely geometric information, one popular line of methods is based on template matching using a collection of clean CAD models either directly~\cite{li2015database,Indoor2012,litany2015asist}, or through extracted features \cite{song2014sliding,avetisyan2019scan2cad}.
More recent methods are based on point cloud deep nets~\cite{qi2017pointnet,zhou2018voxelnet,lang2019pointpillars,shi2018pointrcnn,voteNet}.
In the context of 3D scene understanding, there have also been promising results on semantic and instance segmentation~\cite{yi2018gspn,choy20194d,graham20183d}.
Most relevant to our work are PointRCNN~\cite{shi2018pointrcnn} and Deep Hough Voting (\votenet)~\cite{voteNet} which demonstrated state-of-the-art 3D object detection in outdoor and indoor scenes, respectively. Notably, these results are achieved \emph{without} using the \rgb input. To leverage this additional information, we propose a way to further boost detection performance in this work.  

\paragraph{3D object detection with \rgbd data.}
Depth and color channels both contain useful information that can be useful for 3D object detection. Prior methods for fusing those two modalities broadly fall into three categories: 2D-driven, 3D-driven, and feature concatenation. The first type of method~\cite{lahoud20172d,qi2018frustum,deng2017amodal,xu2018pointfusion} starts with object detecions in the 2D image, which are then used to guide the search space in 3D. By 3D-driven, we refer to methods that first generate region proposals in 3D and then utilize 2D features to make a prediction, such as the Deep Sliding Shapes~\cite{song2016deep}. Recently more works focus on fusing 2D and 3D features earlier in the process such as Multi-modal Voxelnet~\cite{wang2019densefusion}, AVOD~\cite{ku2018joint}, multi-sensor~\cite{liang2018deep} and 3D-SIS~\cite{hou20193d}. However, all these mostly perform fusion through concatenation of 2D features to 3D features. Our proposed method is more closely related to the third type, but differs from it in two important aspects. First, we propose to make explicit use of geometric cues from the 2D detector and lift them to 3D in the form of pseudo 3D votes. Second, we use a multi-tower architecture~\cite{wang2019makes} to balance features from both modalities, instead of simply training on the concatenated features.

\paragraph{Multi-modal fusion in learning.} How to fuse signals from multiple modalities is an open research problem in other areas than 3D object detection.
For example, the main focus of vision and language research is on developing more effective ways to jointly reason over visual data and texts~\cite{fukui2016multimodal,perez2018film,yu2018beyond} for tasks like visual question answering~\cite{antol2015vqa,johnson2017clevr}.
Another active area of research is video+sound~\cite{owens2016visually,gao20192}, where the additional sound track can either provide supervision signal~\cite{owens2016ambient}, or propose interesting tasks to test joint understanding of both streams~\cite{zhao2018sound}.
Targeting at all such tasks, a recent gradient blending approach~\cite{wang2019makes} is proposed to make the multi-modal network more robust (to over-fitting and different convergence rates), which is adopted in our approach too.

\section{ImVoteNet Architecture}
We design a 3D object detection solution suited for \rgbd scenes, based on the recently proposed deep Hough voting framework (\votenet~\cite{voteNet}) by passing \emph{geometric} and \emph{semantic/texture} cues from 2D images to the voting process (as illustrated in Fig.~\ref{fig:pipeline}).
In this section, after a short summary of the original \votenet pipeline, we describe how to build `2D votes' with the assistance of 2D detectors on \rgb and explain how the 2D information is lifted to 3D and passed to the point cloud to improve the 3D voting and proposal. Finally, we describe our multi-tower architecture for fusing 2D and 3D detection with gradient blending~\cite{wang2019makes}. More implementation details are provided in supplement.

\subsection{Deep Hough Voting}
\votenet~\cite{voteNet} is a feed-forward network that consumes a 3D point cloud and outputs object proposals for 3D object detection. Inspired by the seminal work on the generalized Hough transform~\cite{ballard1981generalizing}, \votenet proposes an adaptation of the voting mechanism for object detection to a deep learning framework that is fully differentiable. 

Specifically, it is comprised of a point cloud feature extraction module that enriches a subsampled set of scene points (called \textit{seeds}) with high-dimensional features (bottom of Fig.~\ref{fig:pipeline} from $N{\times}3$ input points to $K{\times}(3{+}F)$ seeds). These features are then pushed through a Multi-Layer-Perceptron (MLP) to generate \emph{votes}. Every vote is both a point in the 3D space with its Euclidean coordinates (3-dim) supervised to be close to the object center, and a feature vector learned for the final detection task (F-dim). The votes form a clustered point cloud near object centers and are then processed by another point cloud network to generate object proposals and classification scores. This process is equivalent to the pipeline in Fig.~\ref{fig:pipeline} with just the point tower and without the image detection and fusion.

\votenet recently achieved state-of-the-art results on indoor 3D object detection in \rgbd~\cite{voteNet}. Yet, it is solely based on point cloud inputs and neglects the image channels which, as we show in this work, are a very useful source of information. In \imvotenet, we leverage the additional image information and propose a lifting module from 2D votes to 3D that improves detection performance. Next, we explain how to get 2D votes in images and how we lift its geometric cues to 3D together with semantic/texture cues.

\subsection{Image Votes from 2D Detection}
\label{sec:image_vote_method}
We generate image votes based on a set of candidate boxes from 2D detectors. An \emph{image vote}, in its geometric part, is simply a vector connecting an image pixel and the center of the 2D object bounding box that pixel belongs to (see Fig.~\ref{fig:teaser}). Each image vote is also augmented with its semantic and texture cues from the features of its source pixel, such that each image vote has $F'$ dimension in total as in the fusion block in Fig.~\ref{fig:pipeline}.

To form the set of boxes given an \rgb image, we apply an off-the-shelf 2D detector (\eg Faster R-CNN~\cite{ren2015faster}) pre-trained on color channels of the \rgbd dataset. The detector outputs the $M$ most confident bounding boxes and their corresponding classes.
We assign each pixel within a detected box a vote to the box center. Pixels inside multiple boxes are given multiple votes (corresponding 3D seed points are duplicated for each of them), and those outside of any box are padded with zeros. Next we go to details on how we derive geometric, semantic and texture cues.

\paragraph{Geometric cues: lifting image votes to 3D}
The translational 2D votes provide useful geometric cues for 3D object localization. Given the camera matrix, the 2D object center in the image plane becomes a ray in 3D space connecting the 3D object center and the camera optical center (Fig.~\ref{fig:teaser}). Adding this information to a seed point can effectively narrow down the 3D search space of the object center to 1D.

\begin{figure}[t]
    \centering
    \includegraphics[width=1.0\linewidth]{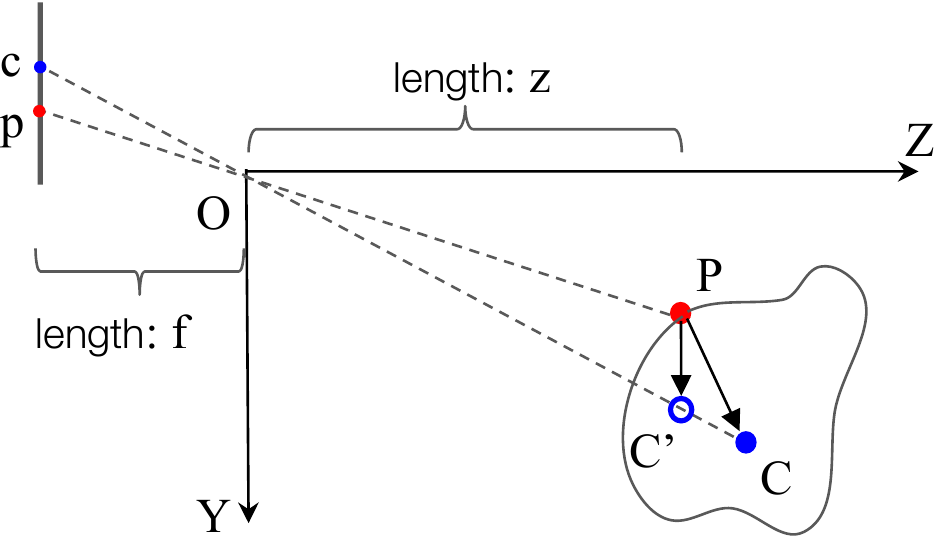}
    \caption{\textbf{Illustration of the pseudo 3D vote.} In the figure, $P$ is a surface point in 3D, $C$ is the unknown object center while $p$ and $c$ are their projections on the image plane respectively. $C'$ is the pseudo 3D center and the vector $\protect\vv{PC'}$ is the pseudo 3D vote.}
    \label{fig:2d_to_3d}
\end{figure}

In details, as shown in Fig.~\ref{fig:2d_to_3d}, given an object in 3D with its detected 2D bounding box in the image plane, we denote the 3D object center as $C$ and its projection onto the image as $c$. A point $P$ on the object surface is associated with its projected point $p$ in the image place, hence knowing the 2D vote to the 2D object center $c$, we can reduces the search space for the 3D center to a 1D position on the ray $OC$. We now derive the computation we follow to pass the ray information to the a 3D seed point. Defining $P{=}(x_1, y_1, z_1)$ in the camera coordinate, and $p{=}(u_1, v_1)$, $c{=}(u_2, v_2)$ in the image plane coordinate, we seek to recover the 3D object center $C{=}(x_2, y_2, z_2)$ (voting target for the 3D point $P$).
The true 3D vote from $P$ to $C$ is:

\begin{equation}
    \vv{PC} = (x_2 - x_1, y_2 - y_1, z_2 - z_1). 
\end{equation}

The 2D vote, assuming a simple pin-hole camera~\footnote{See supplementary for more details on how to deal with a general camera model and camera-to-world transformations.} with focal length $f$, can be written as:
\begin{equation}
\begin{split}
    \vv{pc} & = (u_2 - u_1, v_2 - v_1) = (\Delta u, \Delta v)\\
    & = (f(\frac{x_2}{z_2} - \frac{x_1}{z_1}), f(\frac{y_2}{z_2} - \frac{y_1}{z_1})).
\end{split}
\end{equation}

We further assume the depth of the surface point $P$ is similar to the center point $C$. This is a reasonable assumption for most objects when they are not too close to the camera. 
Then, given $z_1{\approx}z_2$, we compute $\vv{PC'}$, 

\begin{equation}
\begin{split}
    \vv{PC'} = (\frac{\Delta u}{f}z_1, \frac{\Delta v}{f}z_1, 0) ,
\end{split}
\end{equation}
which we refer to as a \emph{pseudo 3D vote}, as $C'$ lies on the ray $OC$ and is in the proximity of $C$.
This pseudo 3D vote provides information about where the 3D center is relative to the point surface point $P$. 

To compensate for the error caused by the depth approximation ($z_1 \approx z_2$), we pass the ray direction as extra information to the 3D surface point. 
The error (along the $X$-axis) caused by the approximated depth, after some derivation, can be expressed by
\begin{equation}
\text{err}_x{=}\Delta x - \Delta x'{=} \frac{x_2}{z_2} (z_2 - z_1).
\end{equation}

Hence, if we input the direction of the ray $\vv{OC}$: $({x_2}/{z_2}, {y_2}/{z_2})$, the network should have more information to estimate the true 3D vote by estimate the depth different $\Delta z = z_2 - z_1$. As we do not know the true 3D object center $C$, we can use the ray direction of $\vv{OC'}$ which aligns with $\vv{OC}$ after all, where
\begin{equation}
\begin{split}
\vv{OC'} & = \vv{OP} + \vv{PC'} \\
& = (x_1 + \frac{\Delta u}{f} z_1, y_1 + \frac{\Delta v}{f} z_1, z_1).
\end{split}
\end{equation}

Normalizing and concatenating with the pseudo vote, the image geometric features we pass to the seed point $P$ are:
\begin{equation}
\label{eq:2dvote}
    (\frac{\Delta u}{f}z_1, \frac{\Delta v}{f}z_1, \frac{\vv{OC'}}{\left\lVert\vv{OC'}\right\rVert}).
\end{equation}

\paragraph{Semantic cues}
On top of the geometric features just discussed that just use the spatial coordinates of the bounding boxes, an important type of information \rgb can provide is features that convey a semantic understanding of what's inside the box. This information often complements what can be learned from 3D point clouds and can help to distinguish between classes that are geometrically very similar (such as table \vs desk or nightstand \vs dresser).

In light of this, we provide additional \emph{region}-level features extracted per bounding box as semantic cues for 3D points. For all the 3D seed points that are projected within the 2D box, we pass a vector representing that box to the point. If a 3D seed point falls into more than one 2D boxes (\ie, when they overlap), we duplicate the seed point for each of the overlapping 2D regions (up to a maximum number of $K$). If a seed point is not projected to any 2D box, we simply pass an all-zero feature vector for padding.

It is important to note that the `region features' here include but are \emph{not} limited to features extracted from \roi pooling operations~\cite{ren2015faster}. In fact, we find representing each box with a simple one-hot class vector (with a confidence score for that class) is already sufficient to cover the semantic information needed for disambiguation in 3D. It not only gives a light-weight input (\eg 10-dim~\cite{sung2015data} \vs 1024-dim~\cite{lin2017feature}) that performs well, but also generalizes to \emph{all} other competitive (\eg faster) 2D detectors~\cite{redmon2016you,liu2016ssd,lin2017focal} that do not explicitly use \roi but directly outputs classification scores. Therefore, we use this semantic cue by default.

\paragraph{Texture cues} 
Different from the depth information that spreads sparsely in the 3D space, \rgb images can capture high-resolution signals at a dense, per-\emph{pixel} level in 2D. While region features can offer a high-level, semantic-rich representation per bounding box, it is complementary and equally important to use the low-level, texture-rich representations as another type of cues. Such cues can be passed to the 3D seed points via a simple mapping: a seed point gets pixel features from the corresponding pixel of its 2D projection\footnote{If the coordinates after projection is fractional, bi-linear interpolation is used.}.

Although any learned, convolutional feature maps with spatial dimensions (height and width) can serve our purpose, by default we still use the simplest texture feature by feeding in the raw \rgb pixel-values directly. Again, this choice is not only light-weight, but also makes our pipeline \emph{independent} of 2D networks. 

Experimentally, we show that even with such minimalist choice of both our semantic and texture cues, significant performance boost over geometric-only \votenet can be achieved with our multi-tower training paradigm, which we discuss next.


\begin{table*}[t!]
\centering
\tablestyle{3pt}{1.2}
\begin{tabular}{l|c|x{30}x{30}x{30}x{30}x{30}x{30}x{30}x{30}x{30}x{30}|x{25}}
 methods & \rgb & bathtub & bed & bookshelf & chair & desk & dresser & nightstand & sofa & table & toilet & mAP \\
\shline
DSS~\cite{song2016deep} & \cmark & 44.2 & 78.8 & 11.9 & 61.2 & 20.5 & 6.4 & 15.4 & 53.5 & 50.3 & 78.9 & 42.1    \\
COG~\cite{ren2016three} & \cmark & 58.3 & 63.7 & 31.8 & 62.2 & {\bf 45.2} & 15.5 & 27.4 & 51.0 & {\bf 51.3} & 70.1 & 47.6 \\
2D-driven~\cite{lahoud20172d} & \cmark & 43.5 & 64.5 & 31.4 & 48.3 & 27.9 & 25.9 & 41.9 & 50.4 & 37.0 & 80.4 & 45.1  \\
PointFusion~\cite{xu2018pointfusion} & \cmark & 37.3 & 68.6 & 37.7 & 55.1 & 17.2 & 23.9 & 32.3 & 53.8 & 31.0 & 83.8 & 45.4 \\ 
F-PointNet~\cite{qi2018frustum} & \cmark & 43.3 & 81.1 & 33.3 & 64.2 & 24.7 & 32.0 & 58.1 & 61.1 & 51.1 & {\bf 90.9} & 54.0 \\ 
\hshline
\votenet~\cite{voteNet} & \xmark & 74.4 & 83.0 & 28.8 & 75.3 & 22.0 & 29.8 & 62.2 & 64.0 & 47.3 & 90.1 & 57.7 \\
\hspace{3mm}+\rgb & \cmark & 70.0 & 82.8 & 27.6 & 73.1 & 23.2 & 27.2 & 60.7 & 63.7 & 48.0 & 86.9 & 56.3 \\
\hspace{3mm}+region feature & \cmark & 71.7 & 86.1 & 34.0 & 74.7 & 26.0 & 34.2 & 64.3 & 66.5 & 49.7 & 88.4 & 59.6 \\
\hshline
\imvotenet & \cmark & {\bf 75.9} & {\bf 87.6} & {\bf 41.3} & {\bf 76.7} & 28.7 & {\bf 41.4} & {\bf 69.9} & {\bf 70.7} & 51.1 & 90.5 & {\bf 63.4} \\
\end{tabular}
\vspace{2mm}
\caption{\textbf{3D object detection results on SUN RGB-D v1 val set.} Evaluation metric is average precision with 3D IoU threshold 0.25 as proposed by~\cite{song2015sun}. Note that both COG~\cite{ren2016three} and 2D-driven~\cite{lahoud20172d} use room layout context to boost performance. The evaluation is on the SUN RGB-D v1 data for fair comparisons. 
}
\label{tab:sunrgbd}
\vspace{-1mm}
\end{table*}


\subsection{Feature Fusion and Multi-tower Training}
With lifted image votes and its corresponding semantic and texture cues ($K\times F'$ in the fusion block in Fig.~\ref{fig:pipeline}) as well as the point cloud features with the seed points $K \times F$, each seed point can generate 3D votes and aggregate them to propose 3D bounding boxes (through a voting and proposal module similar to that in~\cite{voteNet}). 
Yet it takes extra care to optimize the deep network to fully utilize cues from all modalities. As a recent paper~\cite{wang2019makes} mentions, without a careful strategy, multi-modal training can actually result in degraded performance as compared to a single modality training. The reason is that different modalities may learn to solve the task at different rates so, without attention, certain features may dominate the learning and result in over-fitting. In this work, we follow the gradient blending strategy introduced in~\cite{wang2019makes} to weight the gradient for different modality towers (by weighting the loss functions).

In our multi-tower formulation, as shown in Fig.~\ref{fig:pipeline}, we have three towers taking seed points with three sets of features: point cloud features only, image features only and joint features. Each tower has the same target task of detecting 3D objects -- but they each have their separate 3D voting and box proposal network parameters as well as their separate losses. The final training loss is the weighted sum of three detection losses:

\begin{equation}
    L = w_\text{img} L_\text{img} + w_\text{point} L_\text{point} + w_\text{joint} L_\text{joint}.
\end{equation}

Within the image tower, while image features alone cannot localize 3D objects, we have leveraged surface point geometry and camera intrinsic to have pseudo 3D votes that are good approximations to the true 3D votes. So combining this image geometric cue with other semantic/texture cues we can still localize objects in 3D with image features only.

Note that, although the multi-tower structure introduces extra parameters, at inference time we no longer need to compute for the point cloud only and the image only towers -- therefore there is minimal computation overhead.

\section{Experiments}
In this section, we first compare our model with previous state-of-the-art methods on the challenging SUN RGB-D dataset (Sec.~\ref{sec:exp:sota}). Next, we provide visualizations of detection results showing how image information helps boost the 3D recognition (Sec.~\ref{sec:exp:qual}). Then, we present an extensive set of analytical experiments to validate our design choices (Sec.~\ref{sec:exp:analysis}). Finally, we test our method in the conditions of very sparse depth, and demonstrate its robustness (Sec.~\ref{sec:exp:sparse}) in such scenarios.

\subsection{Comparing with State-of-the-art Methods}
\label{sec:exp:sota}
\paragraph{Benchmark dataset.}
\label{sec:exp:benchmark}
We use SUN RGB-D~\cite{silberman2012indoor,janoch2013category,xiao2013sun3d,song2015sun} as our benchmark for evaluation, which is a single-view~\footnote{We do not evaluate on the ScanNet dataset~\cite{dai2017scannet} as in \votenet because ScanNet involves \emph{multiple} 2D views for each reconstructed scene -- thus requires extra handling to merge multi-view features.} \rgbd dataset for 3D scene understanding. It consists of \app10K \rgbd images, with \app5K for training. Each image is annotated with amodal oriented 3D bounding boxes. In total, 37 object categories are annotated. Following standard evaluation protocol~\cite{song2016deep}, we only train and report results on the 10 most common categories. To feed the data to the point cloud backbone network, we convert the depth images to point clouds using the provided camera parameters. The \rgb image is aligned to the depth channel and is used to query corresponding image regions from scene 3D points.
\paragraph{Methods in comparison.}
\label{sec:exp:method}
We compare \imvotenet with previous methods that use both geometry and \rgb. Moreover, since previous state-of-the-art (\votenet~\cite{voteNet}) used only geometric information, to better appreciate the improvement due to our proposed fusion and gradient blending modules we add two more strong baselines by extending the basic \votenet with additional features from image.

Among the previous methods designed for \rgbd, 2D-driven~\cite{lahoud20172d}, PointFusion~\cite{xu2018pointfusion} and F-PointNet~\cite{qi2018frustum} are all cascaded systems that rely on 2D detectors to provide proposals for 3D.
Deep Sliding Shapes~\cite{song2016deep} designs a Faster R-CNN~\cite{ren2015faster} style 3D CNN network to generate 3D proposals from voxel input and then combines 3D and 2D \roi features for box regression and classification. COG~\cite{ren2016three} is a sliding shape based detector using 3D HoG like feature extracted from \rgbd data.

\begin{figure*}[t!]
    \centering
    \includegraphics[width=.95\linewidth]{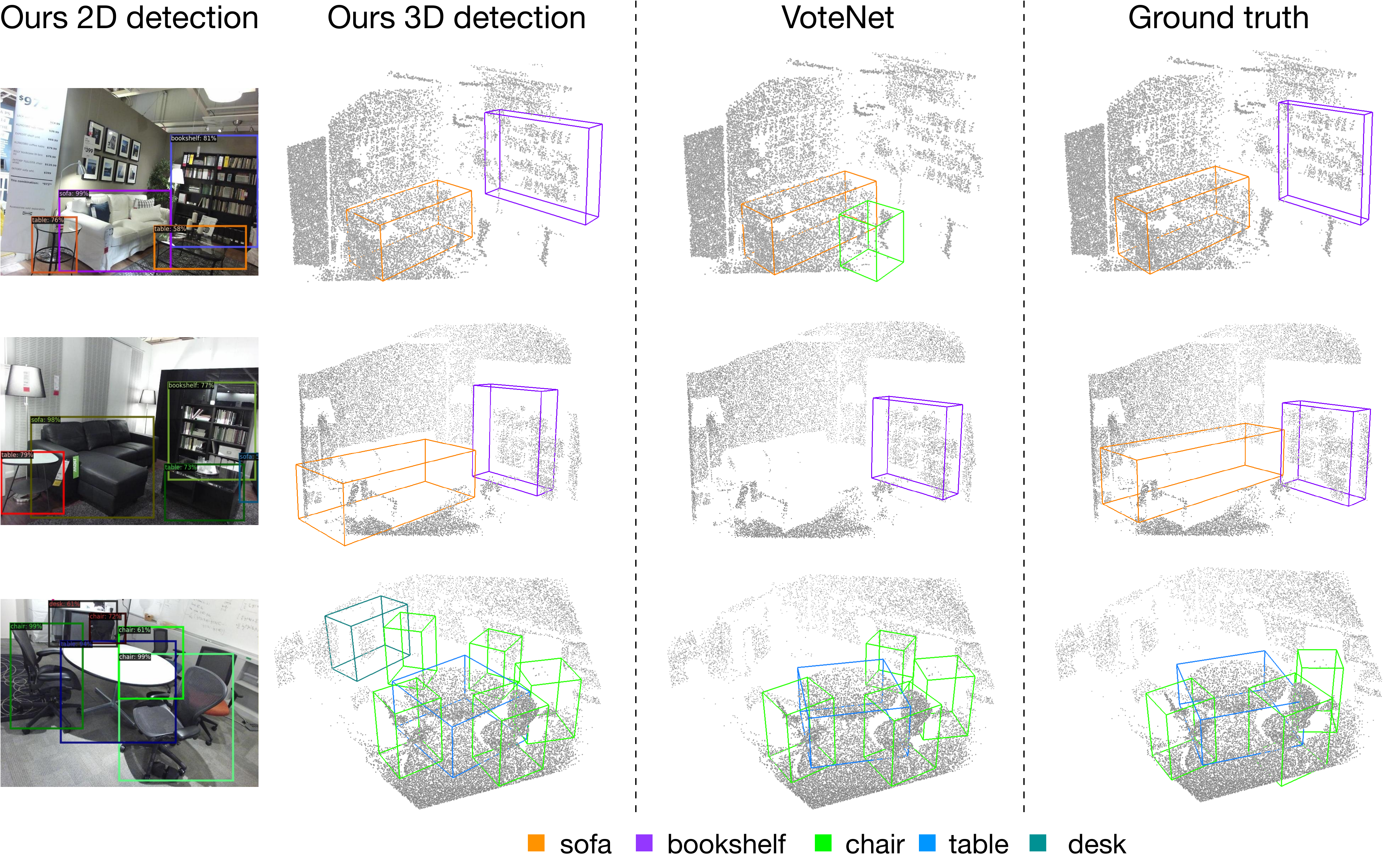}
    \caption{\textbf{Qualitative results showing how image information helps.} First row: the bookshelf is detected by \imvotenet thanks to the cues from the 2D detector; Second row: the black sofa has barely any depth points due to its material, but leveraging images, we can detect it; Third row: with 2D localization cues and semantics, we detect the desk and chairs in the back which are even missed by ground truth annotations. Best viewed in color with zoom in.}
    \label{fig:qualitative}
\end{figure*}

As for the variations of \votenet~\cite{voteNet}, the first one, `+\rgb', directly appends the the \rgb values as a three-dimensional vector to the point cloud features (of the seed points). For the second one (`+region feature'), we use the same pre-trained Faster R-CNN (as in our model) to obtain the region-level one-hot class confidence feature, and concatenate it to the seed points inside that 2D box frustum. These two variations can also be viewed as ablated versions of our method. 

\paragraph{Results.}
Table~\ref{tab:sunrgbd} shows the per-class 3D object detection results on SUN RGB-D. We can see that our model outperforms all previous methods by large margins. Especially, it improves upon the previously best model \votenet by $\textbf{5.7}$ mAP, showing effectiveness of the lifted 2D image votes. It gets better results on nearly all categories and has the biggest improvements on object categories that are often occluded (+12.5 AP for bookshelves) or geometrically similar to the others (+11.6 AP for dressers and +7.7 AP for nightstands).

Compared to the variations of the \votenet that also uses \rgb data, our method also shows significant advantages. Actually we find that naively appending \rgb values to the point features resulted in \emph{worse} performance, likely due to the over-fitting on \rgb values. Adding region features as a one-hot score vector helps a bit but is still inferior compared to our method that more systematically leverage image votes.

\subsection{Qualitative Results and Discussion}
\label{sec:exp:qual}
In Fig.~\ref{fig:qualitative}, we highlight detection results of both the original \votenet~\cite{voteNet} (with only point cloud input) and our \imvotenet with point cloud plus image input, to show how image information can help 3D detection in various ways. The first example shows how 2D object localization and semantic help. We see a cluttered bookshelf that was missed by the \votenet but thanks to the 2D detection in the images, we have enough confidence to recognize it in our network. The image semantics also help our network to avoid the false positive chair as that in the \votenet output (coffee table and candles confused the network there). The second example shows how images can compensate depth sensor limitations. Due to the color and material of the black sofa, there is barely any depth point captured for it. While \votenet completely misses the sofa, our network is able to pick it up. The third example shows how image cues can push the limit of 3D detection performance, by recovering far away objects (the desk and chairs in the back) that are even missed in the ground truth annotations.

\subsection{Analysis Experiments}
\label{sec:exp:analysis}
In this subsection, we show extensive ablation studies on our design choices and discuss how different modules affect the model performance. For all experiments we report mAP@0.25 on SUN RGB-D as before.

\paragraph{Analysis on geometric cues.} To validate that geometric cues lifted from 2D votes help, we ablate geometric features (as in Eq.~\ref{eq:2dvote}) passed to the 3D seed points in Table~\ref{tab:analysis:geometry}. We see that from row 1 to row 3, not using any 2D geometric cue results in a 2.2 point drop. On the other hand, not using the ray angle resulted in a 1.2 point drop, indicating the ray angle helps provide corrective cue to the pseudo 3D votes.

\begin{table*}
\centering
\subfloat[Ablation studies on 2D \textbf{geometric} cues. 2D vote means the lifted 2D vote (2-dim) as in Eq.~\ref{eq:2dvote} and ray angle means the direction of $\protect\vv{OC'}$ (3-dim). Both geometric cues helped our model.
\label{tab:analysis:geometry}]
{\makebox[0.3\linewidth][c]{
\tablestyle{12pt}{1.2}
\begin{tabular}{c|c|c}
\multicolumn{2}{c|}{geometric cues} & \multirow{2}{*}{mAP} \\
\cline{1-2}
2D vote & ray angle & \\
\shline
\cmark&\cmark & 63.4 \\
 \hline
\cmark& \xmark & 62.2 \\
\xmark& \xmark & 61.2 \\
\multicolumn{3}{c}{}\\
\end{tabular}
} 
}
\hfill
\subfloat[Ablation studies on 2D \textbf{semantic} cues. Different region features are experimented. This includes simple one-hot class score vector and rich \roi features. The former (default) works best.
\label{tab:analysis:semantic}]
{\makebox[0.3\linewidth][c]{
\tablestyle{12pt}{1.2}
\begin{tabular}{c|c|c}
\multicolumn{2}{c|}{semantic cues} & \multirow{2}{*}{mAP} \\
\cline{1-2}
region feature & \# dims  & \\
\shline
one-hot score & 10 & 63.4 \\
\hline
\multirow{2}{*}{\roi~\cite{ren2015faster}} & 64 & 62.4 \\
& 1024 & 59.5 \\
\hline
\xmark & - & 58.9 \\
\end{tabular}
}
}
\hfill
\subfloat[Ablation studies on 2D \textbf{texture} cues. We experiment with different pixel-level features including \rgb values (default) and learned representations from the feature pyramid.
\label{tab:analysis:pixel}]
{\makebox[0.3\linewidth][c]{
\tablestyle{12pt}{1.2}
\begin{tabular}{c|c|c}
\multicolumn{2}{c|}{texture cues} & \multirow{2}{*}{mAP} \\
\cline{1-2}
pixel feature & \# dims  & \\
\shline
\rgb & 3 & 63.4 \\
\hline
FPN-$P_2$~\cite{lin2017feature} & 256 & 62.0 \\
FPN-$P_3$ & 256 & 62.0 \\
\hline
\xmark & - & 62.4
\end{tabular}
} 
}
\vspace{1mm}
\caption{{\bf Ablation analysis} on 2D cues. We provide detailed analysis on all types of features from 2D (see Sec.~\ref{sec:image_vote_method} for detailed descriptions).}
\vspace{-1mm}
\end{table*}

\paragraph{Analysis on semantic cues.}
Table~\ref{tab:analysis:semantic} shows how different types of region features from the 2D images affect 3D detection performance. We see that the one-hot class score vector (probability score for the detected class, other classes set to 0), though simple, leads to the best result. Directly using the 1024-dim \roi features from the Faster R-CNN network actually got the worst number likely due to the optimization challenge to fuse this high-dim feature with the rest point features. Reducing the 1024-dim feature to 64-dim helps but is still inferior to the simple one-hot score feature.

\paragraph{Analysis on texture cues.} Table~\ref{tab:analysis:pixel} shows how different low-level image features (texture features) affect the end detection performance. It is clear that the raw \rgb features are already effective while the more sophisticated per-pixel CNN features (from feature pyramids~\cite{lin2017feature} of the Faster R-CNN detector) actually hurts probably due to over-fitting. More details are in the supplementary material.

\paragraph{Gradient blending.} 
Table~\ref{tab:tower_weights} studies how tower weights affect the gradient blending training. We ablate with a few sets of representative weights ranging from single tower training (the first row), dominating weights for each of the tower (2nd to 4th rows) and our best set up. It is interesting to note that even with just the image features (the 1st row, 4th column) i.e. the pseudo votes and semantic/texture cues from the images, we can already outperform several previous methods (see Table~\ref{tab:sunrgbd}), showing the power of our fusion and voting design.

\begin{table}[t]
\tablestyle{10pt}{1.2}
\begin{tabular}{c|c|c|c|c|c}
\multicolumn{3}{c|}{tower weights} & \multicolumn{3}{c}{mAP} \\
\hline
$w_\text{img}$ & $w_\text{point}$ & $w_\text{joint}$ & image & \makecell{point\\cloud} & joint \\
\shline
- & - & - & \unhighlight{46.8} & \unhighlight{57.4} & \unhighlight{62.1} \\
\hline
0.1 & 0.8 & 0.1 & {\bf 46.9} & 57.8 & 62.7 \\
0.8 & 0.1 & 0.1 & 46.8 & {\bf 58.2} & 63.3 \\
0.1 & 0.1 & 0.8 & 46.1 & 56.8 & 62.7 \\
\hline
0.3 & 0.3 & 0.4 & 46.6 & 57.9 & {\bf 63.4} \\
\end{tabular}
\vspace{2mm}
\caption{Analysis on {\bf multi-tower training}. In the first block we show performance \emph{without} blending in gray. Then we show the setting where each of the tower dominates (0.8) the overall training. Finally we show our default setting where weights are more balanced.}
\label{tab:tower_weights}
\vspace{-1mm}
\end{table}

\subsection{Detection with Sparse Point Clouds}
\label{sec:exp:sparse}
While depth images provide dense point clouds for a scene (usually 10k to 100k points), there are other scenarios that only sparse points are available. One example is when the point cloud is computed through visual odometry~\cite{nister2004visual} or Structure from Motion (SfM)~\cite{koenderink1991affine} where 3D point positions are triangulated by estimating poses of a monocular camera in multiple views. With such sparse data, it is valuable to have a system that can still achieve decent detection performance.

To analyze the potential of our model with sparse point clouds, we simulate scans with much less points through two types of point sub-sampling: uniformly random sub-sampling (remove existing points with a uniform distribution) and ORB~\cite{rublee2011orb} key-point based sub-sampling (sample ORB key points on the image and only keep 3D points that project close to those 2D key points). In Table~\ref{tab:sparse}, we present detection results with different distribution and density of point cloud input. We see that in the column of ``point cloud'', with decreased number of points, 3D detection performance quickly drops. On the other hand, we see including image cues significantly improves performance. This improvement is most significant when the sampled points are from ORB key points that are more non-uniformly distributed.

\begin{table}[t]
\tablestyle{12pt}{1.2}
\begin{tabular}{c|c|c|c|c}
\multicolumn{2}{c|}{point cloud settings} & \multicolumn{3}{c}{mAP} \\
\hline
\makecell{sampling\\method} & \# points & \makecell{point\\cloud} & joint & $\Delta$ \\
\shline
\multirow{3}{*}{\makecell{random\\uniform}} & {20k} & {57.7} & {63.4} & \emph{+5.7} \\
\cline{2-5}
 & 5k & 56.2 & 61.7 & \emph{+5.5} \\
 & 1k & 49.6 & 58.5 & \emph{+8.9} \\
\hline
\multirow{2}{*}{ORB~\cite{rublee2011orb}} & 5k & 32.4 & 49.9 & \emph{+16.5} \\
 & 1k & 27.9 & 47.1 & \emph{+19.2} \\
\end{tabular}
\vspace{2mm}
\caption{{\bf Sparse point cloud} experiment, where we sub-sample the number of points in the cloud either via \emph{random uniform} sampling or with \emph{ORB key points}~\cite{rublee2011orb}. In such cases, our \imvotenet significantly outperforms purely geometry based \votenet.}
\label{tab:sparse}
\vspace{-1mm}
\end{table}

\section{Conclusion}
In this work we have explored how image data can assist a voting-based 3D detection pipeline. The \votenet detector we build upon relies on a voting mechanism to effectively aggregate geometric information in point clouds. We have demonstrated that our new network, \imvotenet, can leverage extant image detectors to provide both geometric and semantic/texture information about an object in a format that can be integrated into the 3D voting pipeline. Specifically, we have shown how to lift 2D geometric information to 3D, using knowledge of the camera parameters and pixel depth. \imvotenet significantly boosts 3D object detection performance exploiting multi-modal training with gradient blending, especially in settings when the point cloud is sparse or unfavorably distributed. 


{\small
\bibliographystyle{ieee_fullname}
\bibliography{pcl}
}

\newpage
\clearpage
\appendix
\section*{Supplementary}
\section{Overview}
In this supplementary, we provide more details on the \ours architecture in Sec.~\ref{supp:sec:arch}, including point cloud network architecture, 2D detector, 2D votes and image votes lifting. We also show visualizations of the sparse point clouds in Sec.~\ref{supp:sec:visu}.

\section{Details on \ours Architecture}
\label{supp:sec:arch}
In this section, we explain the details in the \ours architecture. Sec.~\ref{supp:sec:pc} provides details in the point cloud deep net as well as the training procedure. Further details on the 2D detector and 2D votes are described in Sec.~\ref{supp:sec:2d} while details on lifting 2D votes with general camera parameters are described in Sec.~\ref{supp:sec:lift}.

\subsection{Point Cloud Network}
\label{supp:sec:pc}
\paragraph{Input and data augmentation.} The point cloud backbone network takes a randomly sampled point cloud of a SUN RGB-D~\cite{song2015sun} depth image with $20k$ points. Each point has its $XYZ$ coordinate as well as its height (distance to floor). The floor height is estimated as the 1\% percentile of heights of the all points.
Similar to~\cite{voteNet}, we augment the input point cloud by randomly sub-sampling the points from the depth image points on-the-fly. Points are also randomly flipped in both horizontal directions and randomly rotated along the up-axis by Uniform[-30,30] degrees. Points are also randomly scaled by Uniform[-.85, 1.15]. Note that the point height and the camera extrinsic are updated accordingly with the augmentation.
\paragraph{Network architecture.} We adopt the same PointNet++~\cite{qi2017pointnetplusplus} backbone network as that in~\cite{voteNet} with four set abstraction (SA) layers and two feature propagation/upsamplng (FP) layers. With input of $N{\times}4$ where $N{=}20k$, the output of the backbone network is a set of seed points of $K{\times}(3+C)$ where $K{=}1024$ and $C{=}256$.

As for voting, different from \votenet~ that directly predicts votes from the seed points, here we fuse lifted image votes and the seed points before voting. As each seed point can fall into multiple 2D detection boxes, we duplicate a seed point $q$ times if it falls in $q$ overlapping boxes. Each duplicated seed point has its feature augmented with a concatenation of the following image vote features: 5-dim lifted geometric cues (2 for the vote and 3 for the ray angle), 10-dim (per-class) semantic cues and 3-dim texture cues. In the end the fused seed point has 3-dim $XYZ$ coordinate and a 274-dim feature vector.

The voting layer takes the seed point and maps its features to votes through a multi-layer perceptron (MLP) with FC output sizes of 256, 256 and 259, where the
last FC layer outputs XYZ offset and feature residuals (with regard to the 256-dim seed feature) for the votes.
As in~\cite{voteNet}, the proposal module is another set abstraction layer that takes in the generated votes and generate proposals of shape $K' {\times}(5{+}2NH{+}4NS{+}NC)$ where $K'$ is the number of total duplicated seed points and the output dimension consists of 2 objectness scores, 3 center regression values, $2NH$ numbers for heading regression ($NH$ heading bins) and $4NS$ numbers for box size regression ($NS$ box anchors) and $NC$ numbers for semantic classification.

\paragraph{Training procedure.} We pre-train the 2D detector as described more in Sec.~\ref{supp:sec:2d} and use the extracted image votes as extra input to the point cloud network. We train the point cloud deep net with the Adam optimizer with batch size 8 and an initial learning rate of 0.001. The learning rate is decayed by $10\times$ after 80 epochs and then decayed by another $10\times$ after 120 epochs. Finally, the training stops at 140 epochs as we find further training does not improve performance.

\subsection{2D Detector and 2D Cues}
\label{supp:sec:2d}
\paragraph{2D detector training.} While \ours can work with any 2D detector, in this paper we choose Faster R-CNN~\cite{ren2015faster}, which is the current dominant framework for bounding box detection in \rgb. The detector we used has a basic ResNet-50~\cite{he2016deep} backbone with Feature Pyramid Networks (FPN)~\cite{lin2017feature} constructed as $\{P_2, P_3, \ldots, P_6\}$. It is pre-trained on the COCO \emph{train2017} dataset~\cite{lin2014microsoft} achieving a \emph{val2017} AP of 41.0. To adapt the COCO detector to the specific dataset for 2D detection, we further fine-tune the model using ground truth 2D boxes from the training set of SUN-RGBD before applying the model only using the color channels. The fine-tuning lasts for 4K iterations, with the learning rate reduced by 10${\x}$ at 3K-th iteration starting from 0.01. The batch size, weight decay, and momentum are set as 8, 1e-4, and 0.9, respectively. Two data augmentation techniques are used: 1) standard left-right flipping; and 2) scale augmentation by randomly sample the shorter side of the input image from [480,600]. The resulting detector achieves a mAP (at overlap 0.5) of 58.5 on val set.

Note that we specifically choose \emph{not} to use the most advanced 2D detectors (\eg based on ResNet-152~\cite{he2016deep}) just for the sake of performance improvement. As our experimental results shown in the main paper, even with this simple baseline Faster R-CNN, we can already see significant boost thanks to the design of \ours.

\paragraph{2D boxes.} To infer 2D boxes using the detector, we first resize the input image to a shorter side of 600 before feeding into the model. Then top 100 detection boxes across all classes for an image is aggregated. We further reduce the number of 2D boxes per image by filtering out any detection with a confidence score below 0.1. Two things to note about the 2D boxes used while training \ours: 1) we could also train with ground truth 2D boxes, however we empirically found that including them for training hurts performance, likely due to the different detection statistics at test time; 2) as the pre-training for the 2D detector is also performed on the same training set, it generally gives better detection results on SUN RGB-D train set images, to reduce the effect of over-fitting, we randomly dropped 2D boxes with a probability of 0.5. 

\paragraph{Alternative semantic cues.} Other than the default semantic cue to represent each 2D box region as the one-hot classification score vector (the detected class has the value of the confidence score from the detector, all other locations have zeros), we further experimented with dense \roi features extracted from that region. Two variants are reported in the paper, with the 1024-dim one being the output from the last FC layer \emph{before} region classification and regression. For the 64-dim one, we insert an additional FC layer before the final output layers so that region information is compressed into the 64-dim vector. The added layer is pre-trained with the 2D detector (resulting in a val mAP of 57.9) and fixed when training \ours. 

\paragraph{Alternative texture cues.} The default texture cue is the raw \rgb values (normalized to [-1,1]). Besides this simple texture cue, we also experimented more advanced per-pixel features. One handy feature that preserves such spatial information is the feature maps $P_k$ from FPN that fuse top-down and lateral connections~\cite{lin2017feature}. Here $k$ is the index to the layers in the feature pyramid, which also designates the feature strides and size. For example, $P_2$ has a stride of 2$^2$=4 for both height and width; and a spatial size of roughly 1$/$16 of the input image\footnote{Note that different from 2D box detection, we feed the images directly to the model \emph{without} resizing to shorter-side 600 to compute FPN features.}. For $P_3$ the strides are 2$^3$=8. All feature maps have a channel size of 256, which becomes the input dimension when used as texture cues for \ours. 

\subsection{Image Votes Lifting}
\label{supp:sec:lift}

\begin{figure}[t!]
    \centering
    \includegraphics[width=\linewidth]{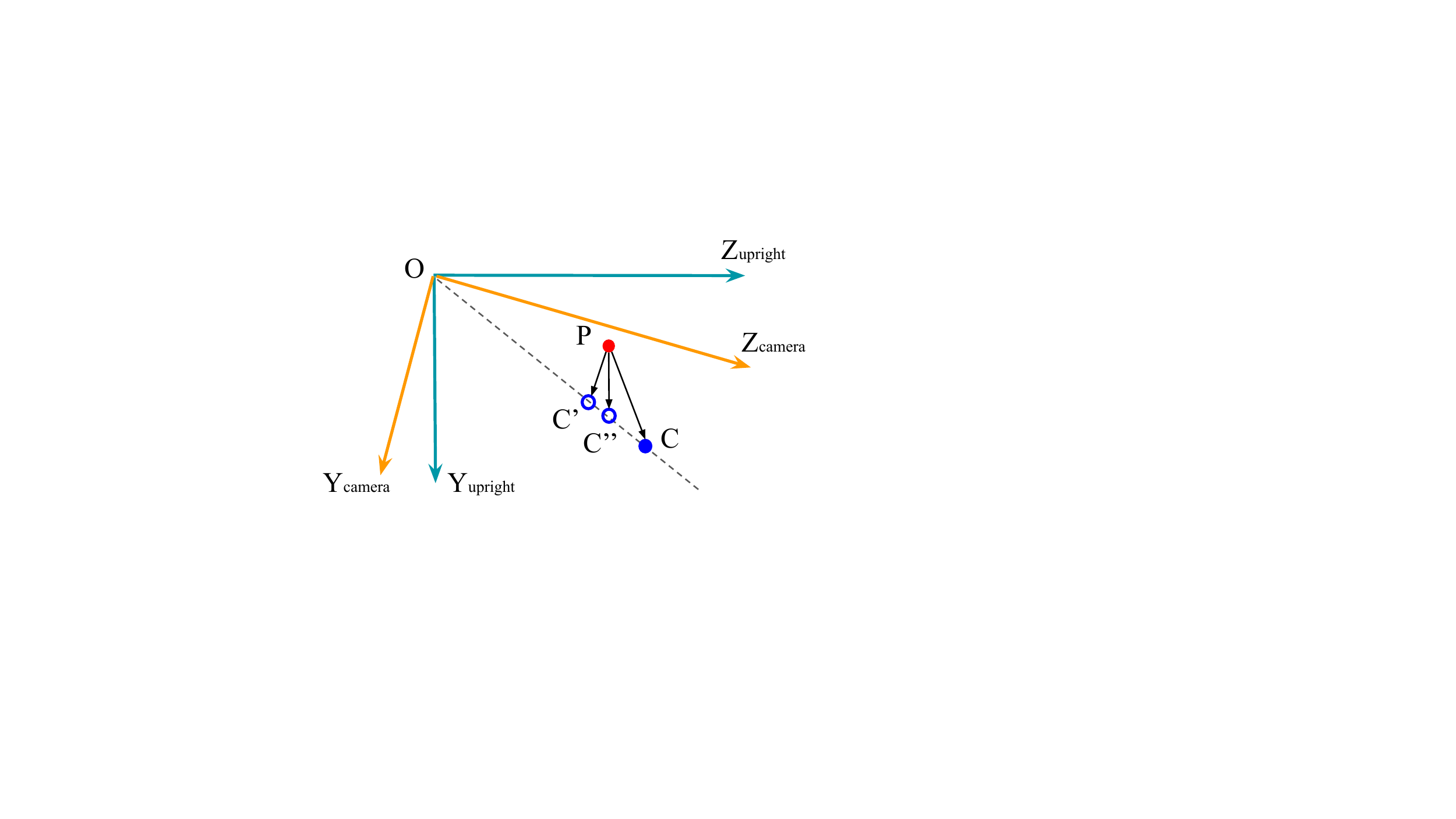}
    \caption{\textbf{Image vote lifting with camera extrinsic.} Here we show surface point $P$ and object center $C$ in two coordinate systems: camera coordinate and upright coordinate ($OY$ is along gravitational direction). $\protect\vv{PC}$ is the true 3D vote. $\protect\vv{PC'}$ is the pseudo vote as calculated in the main paper and $\protect\vv{PC''}$ is the transformed pseudo vote finally used in feature fusion.}
    \label{fig:lift_supp}
\end{figure}

In the main paper we derived the lifting process to transform a 2D image vote to a 3D pseudo vote \emph{without} considering the camera extrinsic. As the point cloud sampled from depth image points is transformed to the \emph{upright coordinate} before feeding to the point cloud network (through camera extrinsic $R$ as a rotational matrix), the 3D pseudo vote also needs to be transformed to the same coordinate.

Fig.~\ref{fig:lift_supp} shows the surface point $P$, object center $C$ and the end point of the pseudo vote $C'$. Since the point cloud is in the upright coordinate, the point cloud deep net can only estimate the depth displacement of $P$ and $C$ along the $Z_{\text{upright}}$ direction (it cannot estimate the depth displacement along the $Z_{\text{camera}}$ direction as the rotational angles from camera to upright coordinate are unknown to the network). Therefore, we need to calculate a new pseudo vote $\vv{PC''}$ where $C''$ is on the ray $OC$ and $PC''$ is perpendicular to the $OZ_{\text{upright}}$.

To calculate the $C''$ we need to firstly transform $P$ and $C'$ to the upright coordinate. Then assume $P{=}(x_p, y_p, z_p)$ and $C'{=}(x_{c'}, y_{c'}, z_{c'})$ in the upright coordinate, we can compute:

\begin{equation}
    C'' = (z_p \frac{x_{c'}}{z_{c'}}, z_p \frac{y_{c'}}{y_{c'}}, z_p).
\end{equation}

\begin{table*}[!t]
\tablestyle{6pt}{1.2}
\begin{tabular}{c|c|c|c|c}
\multicolumn{2}{c|}{point cloud settings} & \multicolumn{3}{c}{images from SUN RGB-D~\cite{song2015sun} train set} \\
\hline
\makecell{sampling\\method} & \# points & 1 & 2 & 3 \\
\shline
\multirow{4}{*}{\makecell{random\\uniform}} & 20k & \raisebox{-0.015\height}{\includegraphics[width=.25\linewidth]{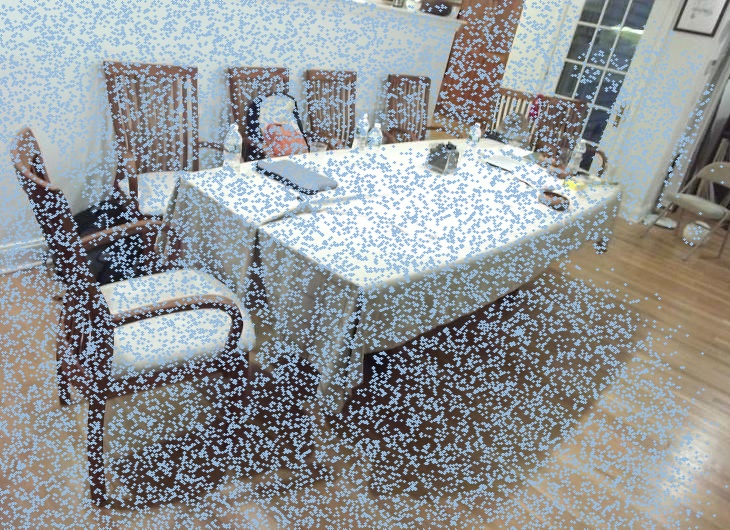}} & \raisebox{-0.015\height}{\includegraphics[width=.25\linewidth]{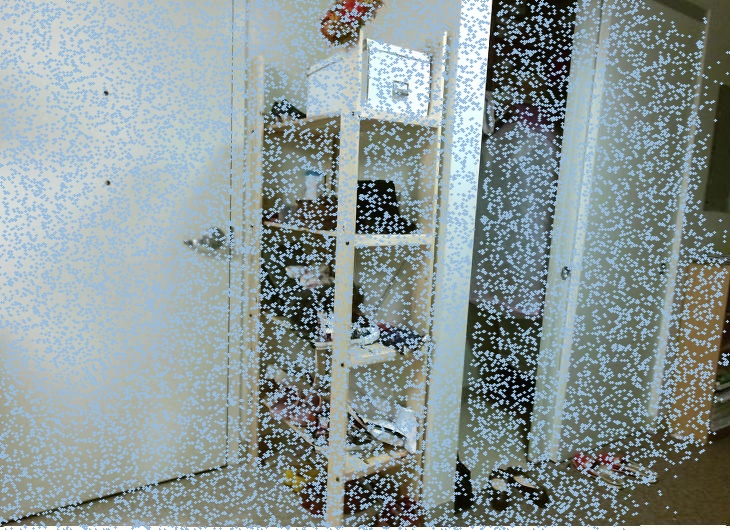}} & \raisebox{-0.015\height}{\includegraphics[width=.25\linewidth]{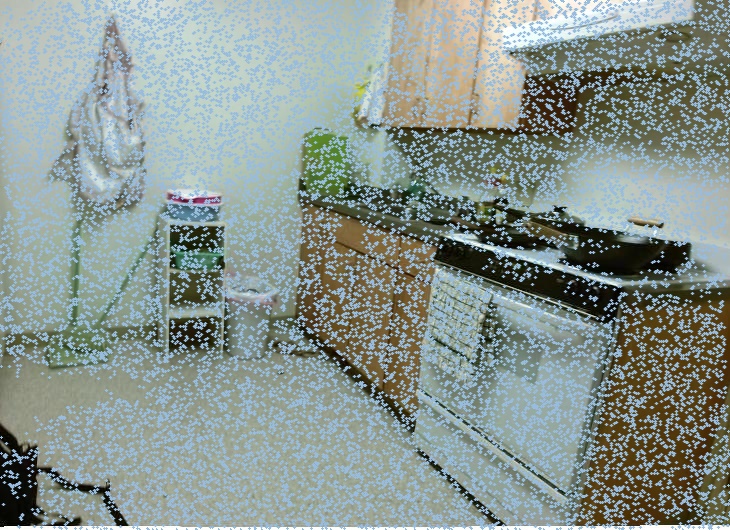}} \\
\cline{2-5}
 & 5k & \raisebox{-0.015\height}{\includegraphics[width=.25\linewidth]{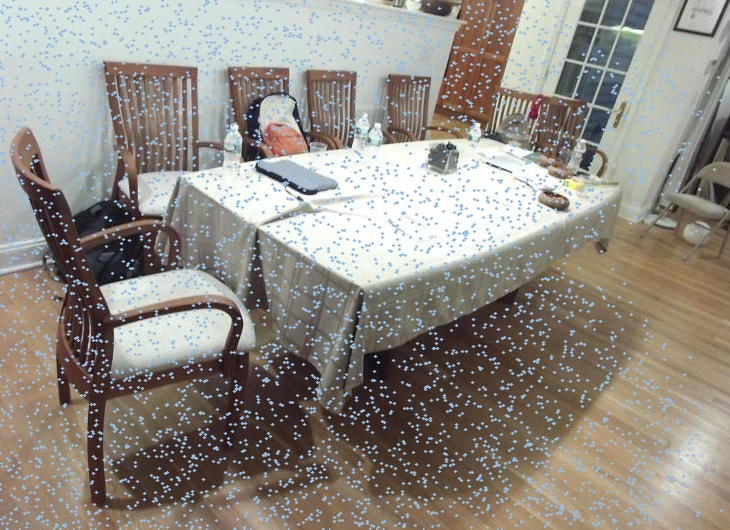}} & \raisebox{-0.015\height}{\includegraphics[width=.25\linewidth]{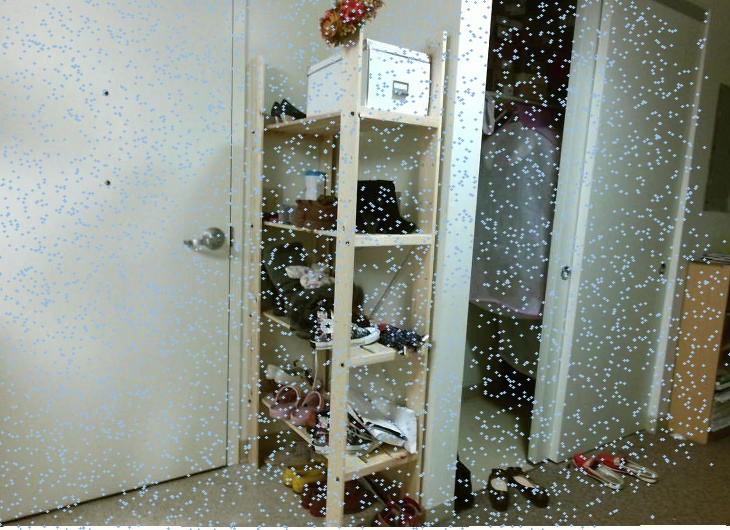}} & \raisebox{-0.015\height}{\includegraphics[width=.25\linewidth]{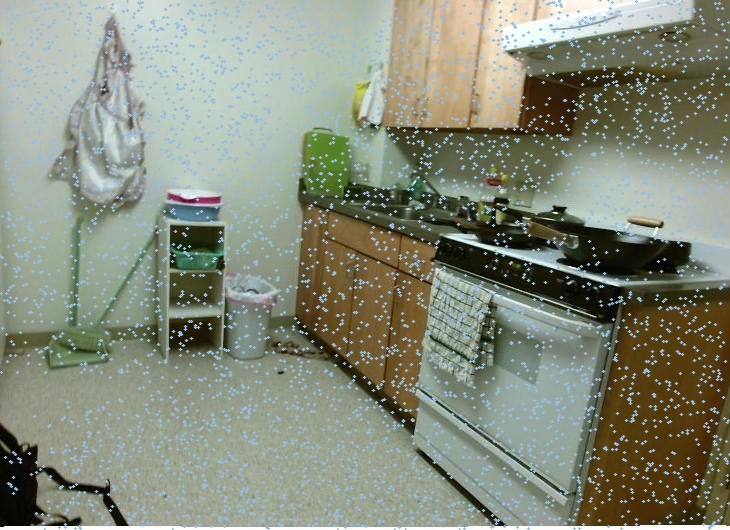}}  \\
 & 1k &  \raisebox{-0.015\height}{\includegraphics[width=.25\linewidth]{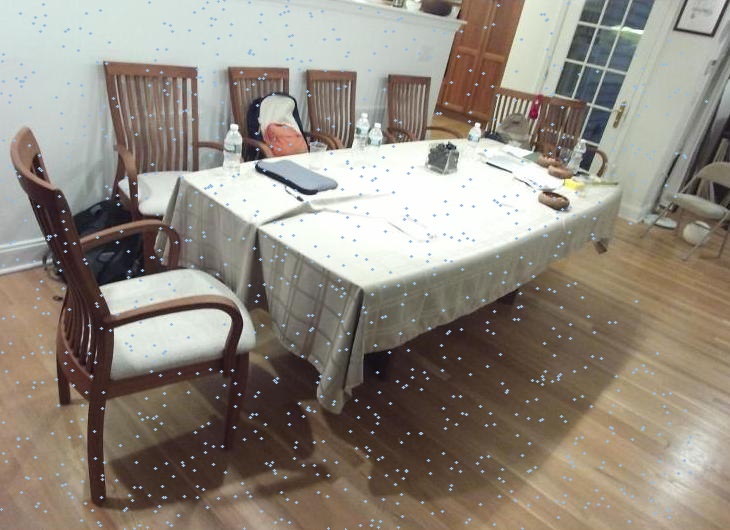}} & \raisebox{-0.015\height}{\includegraphics[width=.25\linewidth]{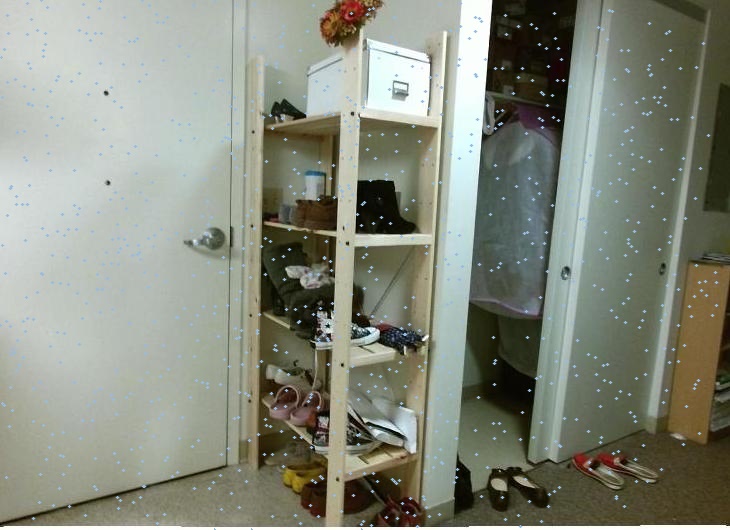}} & \raisebox{-0.015\height}{\includegraphics[width=.25\linewidth]{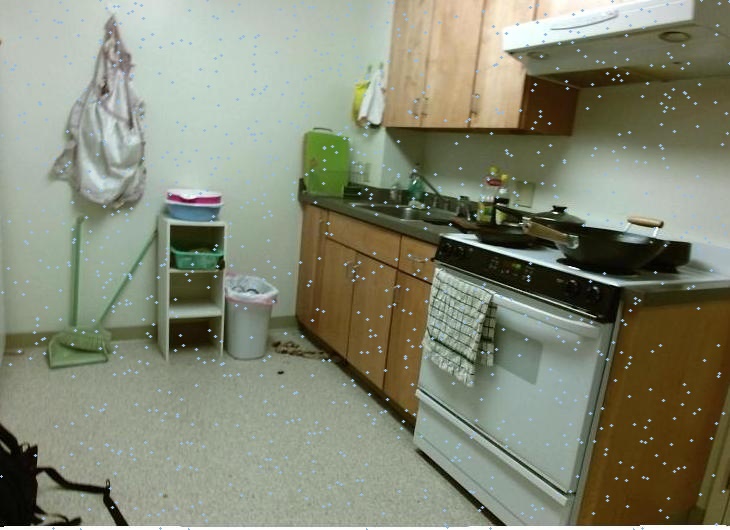}}  \\
\hline
\multirow{2}{*}{ORB~\cite{rublee2011orb}} & 5k & \raisebox{-0.015\height}{\includegraphics[width=.25\linewidth]{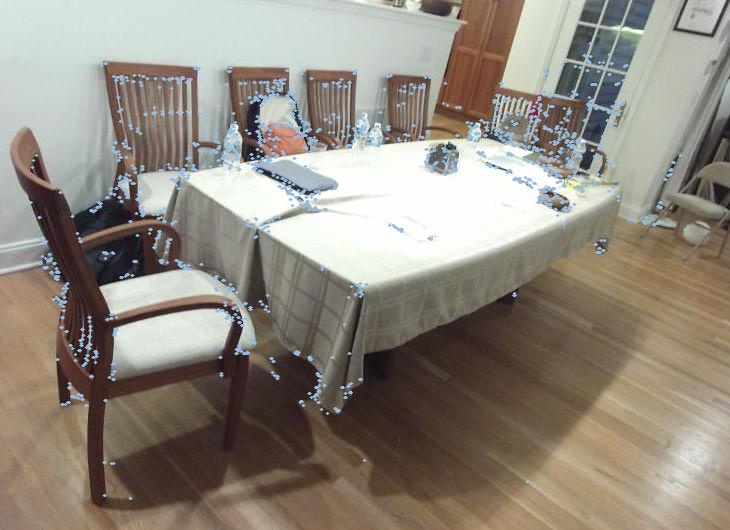}} & \raisebox{-0.015\height}{\includegraphics[width=.25\linewidth]{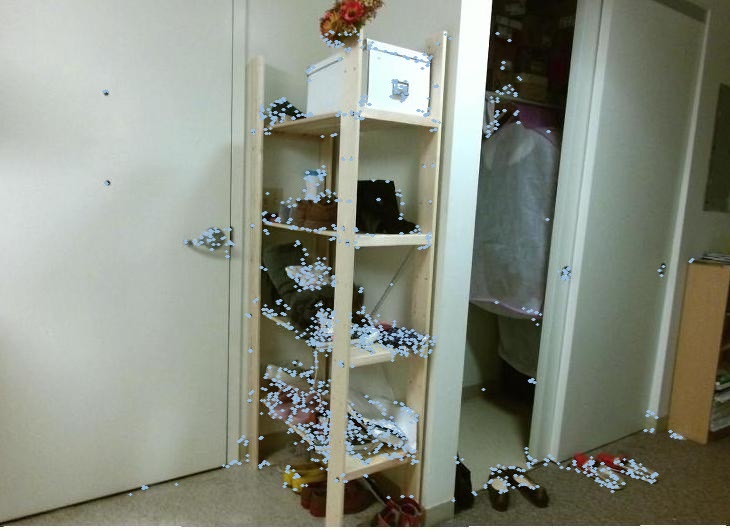}} & \raisebox{-0.015\height}{\includegraphics[width=.25\linewidth]{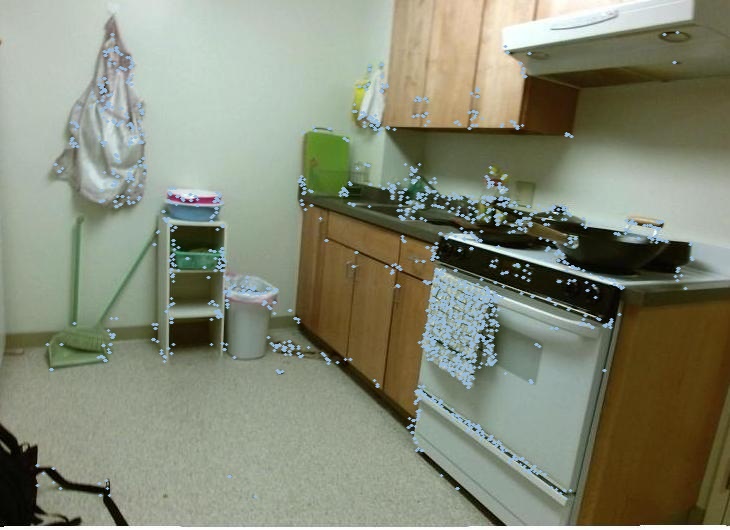}}  \\
 & 1k & \raisebox{-0.015\height}{\includegraphics[width=.25\linewidth]{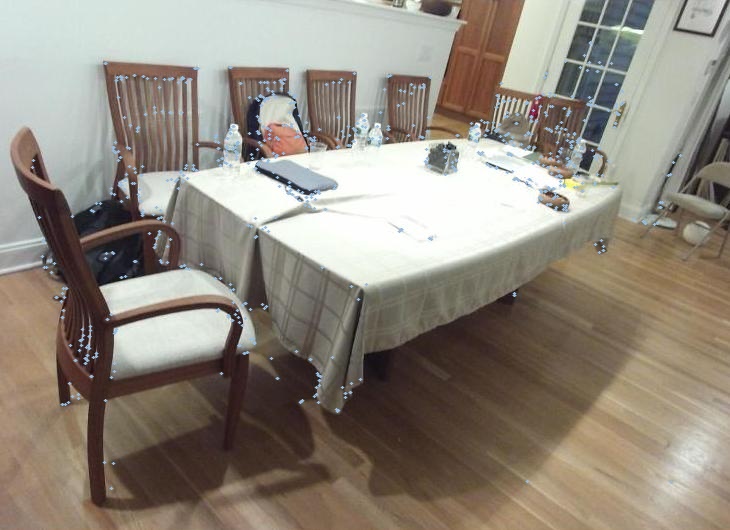}} & \raisebox{-0.015\height}{\includegraphics[width=.25\linewidth]{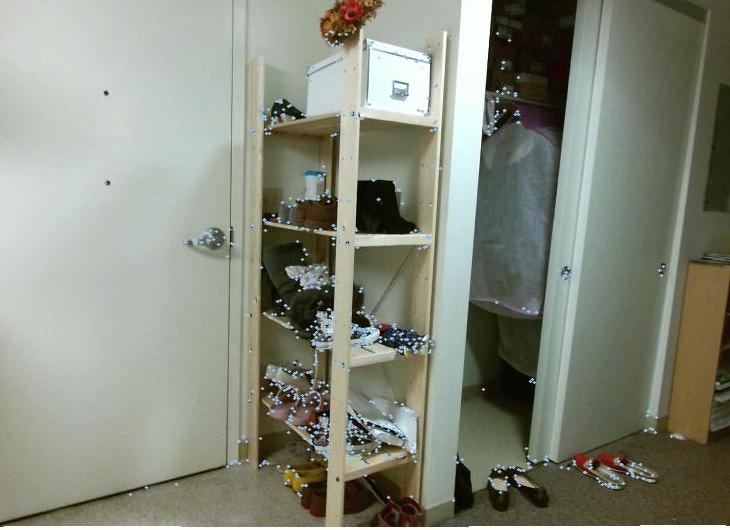}} & \raisebox{-0.015\height}{\includegraphics[width=.25\linewidth]{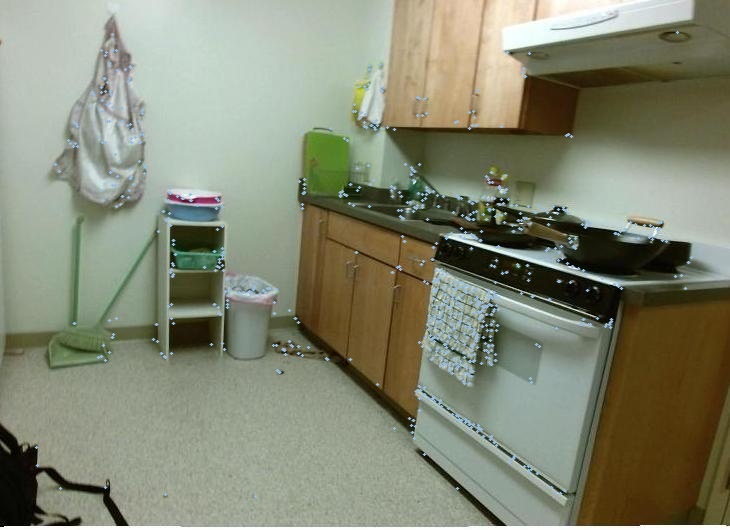}} \\
 \hline
\end{tabular}
\vspace{2mm}
\caption{{\bf Sparse point cloud visualization.} We show projected point clouds on three SUN RGB-D images and compare point density and distribution among random sampling (to $20k$, $5k$ and $1k$ points) and ORB key points based sampling (to $5k$ and $1k$ points). For ORB key point sampling, we firstly detect ORB key points in the RGB images with an ORB key point detector and then keep 3D points that are projected near those key points. Best viewed in color with zoom in. }
\label{tab:sparse}
\vspace{-1mm}
\end{table*}

\section{Visualization of Sparse Points}
\label{supp:sec:visu}
In the Sec. 4.4 of the main paper we showed how image information and \ours model can be specially helpful in detections with sparse point clouds. Here in Table.~\ref{tab:sparse} we visualize the sampled sparse point clouds on three example SUN RGB-D images. We project the sampled points to the RGB images to show their distribution and density. We see that the $20k$ points in the first row have a dense and uniform coverage of the entire scene. After randomly subsampling the points to $5k$ and $1k$ points in the second and third rows respectively, we see the coverage is much more sparse but still uniform. In contrast the ORB key point based sampling (the last two rows) results in very uneven distribution of points where they are clustered around corners and edges. The non-uniformity and low coverage of ORB key points makes it especially difficult to recognize objects in point cloud only. That's also where our \ours model showed the most significant improvement upon \votenet.


\end{document}